\documentclass[conference]{IEEEtran}
\IEEEoverridecommandlockouts
\usepackage{times}
\usepackage{cite}

\usepackage[numbers,compress]{natbib} 
\usepackage[bookmarks=true,colorlinks=true,linkcolor=blue,urlcolor=blue]{hyperref}
\usepackage{multicol}
\usepackage{graphicx}
\usepackage[dvipsnames]{xcolor}
\usepackage{soul}
\usepackage[most]{tcolorbox}
\usepackage{amsmath, amsthm, amsfonts, amssymb, mathtools} 
\usepackage{algorithm}
\usepackage{algpseudocode}
\usepackage{multirow}
\usepackage{svg}
\usepackage{booktabs, makecell}
\hypersetup{
    colorlinks=true,
    linkcolor=blue,      
    urlcolor=blue,
    pdftitle={Overleaf Example},
    pdfpagemode=FullScreen,
}
\usepackage{etoolbox}
\makeatletter
\patchcmd{\@makecaption}
  {\scshape}
  {}
  {}
  {}
\makeatother
 
\DeclareMathOperator*{\argmax}{arg\,max}  

\title{Motion-Uncertainty-Aware Next-Best-View Planning for Moving Object Reconstruction}

\author{Karen Li$^{1}$, Mattia Mantovani$^{2}$, Robert J. Wood$^{1}$, Lorenzo Sabattini$^{2}$, and Stephanie Gil$^{1}$
 
\thanks{$^{1}$Harvard University, Cambridge, MA 02134, USA. 
{\{\tt\small karenli@g, rjwood@seas, sgil@seas\}.harvard.edu}}%
\thanks{$^{2}$University of Modena and Reggio Emilia, 42122 Reggio Emilia,
Italy.
{\{\tt\small name.surname\}@unimore.it}}
{\thanks{*We use target and object interchangeably.}}
{\thanks{Code can be found at \href{https://github.com/Harvard-REACT/mua-nbv}{https://github.com/Harvard-REACT/mua-nbv}}}
}  

\begin{document}
\maketitle


\begin{abstract}  
Active 3D reconstruction of moving objects requires selecting informative viewpoints while accounting for object motion uncertainty during the decision-to-execution delay.
Existing methods address only parts of this problem: next-best-view (NBV) planners for object reconstruction typically optimize surface coverage but assume static objects, while motion-aware active perception for moving targets accounts for target motion but prioritizes tracking or visibility over reconstruction coverage.
This work presents a motion-uncertainty-aware NBV framework for reconstructing an unknown rigid object undergoing planar motion, using noisy planar position measurements of the object and depth observations from a mobile robot.
The key idea is to evaluate each candidate viewpoint by its expected observation quality over plausible future object states induced by motion and measurement uncertainty, rather than at a single predicted object pose.
To obtain this predictive belief, a fixed-lag Gaussian Process smoother estimates and predicts the object state from noisy position measurements.
The resulting belief is used to generate candidate viewpoints around the predicted object location, filter them by reachability, and estimate their expected coverage-driven scores.
Simulation and real-world experiments demonstrate improved reconstruction completeness over non-predictive NBV and prediction-only tracking methods, bridging coverage-driven active reconstruction and prediction-driven tracking.
\end{abstract} 


\section{Introduction} 
{
Active object reconstruction seeks to build a 3D model of an unknown object by selecting informative sensing viewpoints that maximize reconstruction progress, commonly measured by surface coverage~\cite{Pan22,Yan23,Border24}. 
Most next-best-view (NBV) methods in this setting assume static objects~\cite{Pan22,Yan23,Border24,Vasquez14,Monica18,Jia25}, so a viewpoint evaluated during planning is expected to remain informative when executed.
This assumption no longer holds for moving objects: a viewpoint that appears informative during planning can become redundant or poorly aligned by execution time as object motion changes self-occlusion and visible surface regions. 
Thus, active reconstruction of moving objects requires reasoning not only about which viewpoints are informative, but also about how \emph{uncertainty} in the object's future state affects the camera-object configuration at the time of execution. 
 
This problem frequently arises in field robotics, where robots must collect informative visual data from moving targets during opportunistic encounters~\cite{Koger23}. 
Whale research, for example, uses visual observations during surfacing events to support health assessment and long-term monitoring~\cite{Gero14,hirtle22,christiansen19,bierlich24,napoli24}. 
While aerial top-down imaging can estimate quantities such as body length or mass~\cite{christiansen19,bierlich24,napoli24}, building more complete individual-level 3D models requires broader surface coverage than a limited set of views can provide. 
This motivates an active perception problem: given coarse target-position measurements from an overhead sensing platform such as a drone~\cite{Bhattacharya25}, a sensing robot must \emph{actively diversify viewpoints} to observe previously unseen surfaces as the target moves.
}

\begin{figure}[t]
    \centering
    \includegraphics[width=1\linewidth, trim=5 55 0 -20]
    {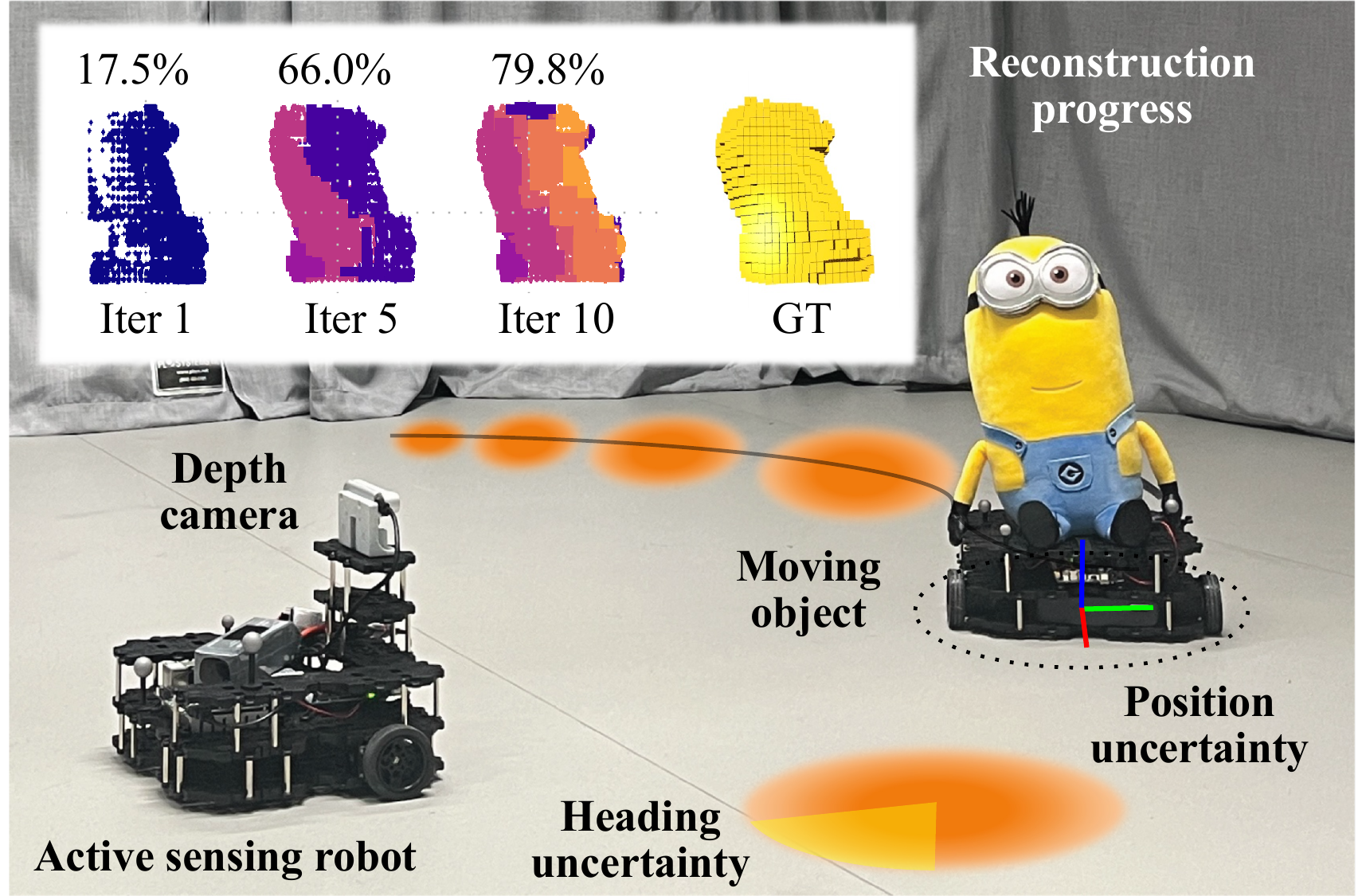} 
    \caption{Overview of the motion-uncertainty-aware NBV planning platform.
    A mobile robot equipped with a depth camera streams observations to an online offboard planner, which selects reachable viewpoints for reconstructing a rigid object undergoing planar motion.
    Orange ellipses indicate noisy object position measurements and predictive object-state uncertainty. 
    The inset shows reconstruction progress over planning iterations, with voxel completeness increasing as new viewpoints observe previously unseen object surfaces.}
    \label{fig:platform}
    \vspace{-0.20 in}
\end{figure}
 
This need for diverse viewpoints under target motion is not fully addressed by existing work.
Prior NBV methods~\cite{Pan22,Yan23,Border24,Vasquez14,Monica18,Jia25} develop coverage-driven objectives for static object reconstruction, while moving-target active perception~\cite{Kiciroglu20,Wang21,Penin18,Tallamraju19} typically focuses on tracking or visibility maintenance. 
To address this gap, this work studies online coverage-driven NBV planning for reconstructing an unknown rigid object undergoing planar motion in an open scene, treating the robot pose as known for planning and modeling uncertainty only in the object state over the planning-to-execution interval. 
This setting is challenging for three reasons:
(i) viewpoint quality depends on the future camera-object configuration, so object motion can change self-occlusion and visible surface regions;
(ii) coverage-driven viewpoint scores depend on nonlinear visibility and discretized surface representations~\cite{Jia25,Delmerico18,Guedon22}, so their expectation under object-state uncertainty does not admit a closed-form expression; and
(iii) the most informative viewpoint under the predictive belief may be unreachable within one planning step due to the robot's kinematic limits. 

Our key idea is to reformulate coverage-driven NBV planning for moving object reconstruction as a belief-space viewpoint selection problem.
Rather than scoring each candidate at a single object pose, candidate viewpoints are evaluated using their expected coverage-driven score under a predictive belief over the execution-time object state.
This belief is obtained by modeling the object trajectory with a Gaussian Process (GP) prior and applying a fixed-lag GP smoother~\cite{Barfoot14,Barfoot13,Tong13} to noisy position-only measurements, yielding a Gaussian estimate of the object's planar position and latent velocity.
Propagating the belief one step forward gives the predictive distribution used for uncertainty-aware candidate generation around the predicted object location, followed by single-step reachability filtering.
Since the expected coverage-driven score has no closed form, each reachable candidate's score is approximated using Monte Carlo samples from the predictive belief, and the highest-scoring viewpoint is selected.
 
The proposed framework is evaluated against two baselines representing the paradigms it bridges: \emph{non-predictive NBV}, which omits one-step object-state prediction, and \emph{prediction-only tracking}, which follows the predicted object location without coverage-driven scoring. 
In both simulation and real-world experiments, the proposed framework improves reconstruction completeness by selecting viewpoints with high expected observation quality under the predicted object-state belief.
The improvement over non-predictive NBV shows the value of accounting for object motion during the planning-to-execution interval, while the improvement over prediction-only tracking shows that following the predicted target is insufficient without coverage-driven viewpoint evaluation.
 
Our main contributions are multifold:
\begin{itemize}
    \item A motion-uncertainty-aware NBV formulation for moving object reconstruction that integrates GP-based object-state prediction with coverage-driven viewpoint selection under execution-time object-state uncertainty. 
    \item An online receding-horizon planning pipeline that uses the predictive belief for uncertainty-aware candidate generation, reachability filtering, and Monte Carlo estimation of the expected coverage-driven viewpoint score.
    \item Simulation and real-world validation showing improved reconstruction completeness over baseline methods, with ablations isolating the effects of candidate generation and viewpoint evaluation.
\end{itemize} 


\section{Related Work}   
Our problem lies at the intersection of two lines of work:
(i) online NBV planning for model-free object reconstruction, and
(ii) predictive perception and view planning for moving targets.
We bridge these lines by formulating moving object NBV planning as belief-space viewpoint selection, where a predictive belief over the execution-time object state is used to evaluate expected viewpoint quality.
The distinctive challenge is that object motion during the planning-to-execution interval makes the execution-time camera-object configuration uncertain, so the planner must reason about how object-state uncertainty \emph{propagates} into viewpoint quality.

\subsection{NBV for Model-Free Object Reconstruction}

NBV planning has a long history in model-free object reconstruction, where a robot incrementally builds an object representation and selects viewpoints that improve reconstruction completeness or surface coverage~\cite{Pan22,Yan23,Border24,Monica18,Vasquez14}. 
Many methods in this setting maintain a voxel-based object representation and evaluate frontier or visibility cues at the boundary between observed and unobserved regions, since this provides a natural representation of reconstruction progress~\cite{Pan22,Delmerico18}.
A direct way to score candidate viewpoints is to ray-cast through this map and count visible frontier, occupied, occluded, or unknown voxels as proxies for expected reconstruction gain~\cite{Santos20,Batinovic21,Delmerico18}.
 
These approaches are effective for static objects, where the camera-object configuration used for scoring remains valid at execution time.
For moving objects, however, object motion can change self-occlusion and surface visibility between planning and execution, making a score computed at a single object pose unreliable.
Thus, we retain the model-free, object-centric reconstruction representation, but evaluate viewpoints under a predictive distribution over the future object state.

Direct voxel-level scoring can be costly for online planning when many viewpoints must be evaluated repeatedly, especially under multiple samples of a predictive belief. 
We therefore instantiate the deterministic per-view score using PB-NBV~\cite{Jia25}, a projection-based surrogate for efficient coverage-driven viewpoint evaluation.
PB-NBV is used as an existing deterministic score; our contribution is to evaluate such scores in expectation under a predictive object-state belief.

\subsection{State Estimation, Prediction via Sparse Gaussian Processes}
 
The belief-space NBV formulation above requires a predictive distribution over the object state at viewpoint execution time.
In our setting, only noisy planar object-position measurements are available, while velocity is latent and is used to define object heading for viewpoint selection.
Thus, the planner must infer both latent velocity and predictive uncertainty from recent observations.
Filtering and smoothing methods are commonly used for this type of state-estimation problem~\cite{pratissoli25,Sahl24,mantovani24}, providing state estimates and predictive uncertainty from noisy partial measurements.
We use a fixed-lag smoother to estimate position and velocity jointly over a short measurement window, a common strategy for online state estimation~\cite{Lagerblad20,Goudar23}. 

Because the planner needs not only a point prediction but also uncertainty over the object's execution-time state, we model the object trajectory probabilistically.
Sparse GP trajectory priors~\cite{Barfoot13,Barfoot14,Tong13,Yan17} provide a lightweight continuous-time model for this short-horizon prediction problem.
Continuous-time trajectory estimation supports prediction at arbitrary execution times within a replanning loop~\cite{Li19}.
These methods exploit the Markov structure induced by stochastic differential equation priors for efficient inference and uncertainty queries in continuous time, making them attractive for bounded online prediction.
We use this machinery in fixed-lag form as an enabling prediction module, where the resulting predictive belief over object state is then consumed by the NBV planner.
 
\subsection{View Planning for Moving Targets}

For moving targets, many active perception systems~\cite{Kiciroglu20,Wang21,Penin18,Tallamraju19} select viewpoints to maintain visibility or improve observability rather than maximize reconstruction coverage.
For example, ActiveMoCap~\cite{Kiciroglu20} selects viewpoints to improve 3D human pose and motion-capture accuracy from a moving camera.
Aerial tracking systems~\cite{Wang21,Saini22,Ho21} similarly plan viewpoints to maintain observability of moving targets.
Visibility-aware tracking methods also plan under target-state uncertainty in cluttered environments~\cite{Gao23}, but the objective remains target tracking.
These methods account for target motion or state uncertainty, but they do not optimize viewpoints for coverage-driven reconstruction of previously unseen object surfaces.

Closer to our setting, prior work on moving object 3D reconstruction considers object motion during viewpoint selection~\cite{zhang2023}.
However, viewpoint quality is evaluated at a single nominal object pose, which is fragile when delay and noisy measurements induce uncertainty over the execution-time object state.
In contrast, our proposed motion-uncertainty-aware NBV framework uses a predictive object-state belief to evaluate coverage-driven viewpoint quality, explicitly accounting for object-state uncertainty induced by object motion.

\begin{figure}
    \centering
    \includegraphics[width=1\linewidth, trim=0 18 0 5]{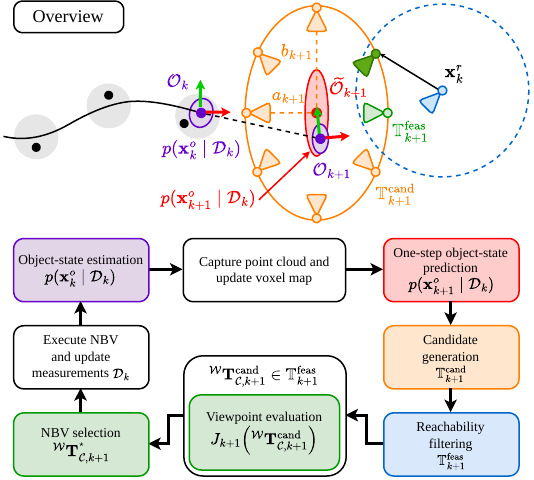}
    \caption{\textbf{Motion-uncertainty-aware NBV online planning loop.} 
    Given noisy object-position measurements (black dots), the GP-based smoother computes the 
    \textcolor{Plum}{\textbf{current belief}} and propagates it forward to the \textcolor{Red}{\textbf{predictive belief}}.
    The \textcolor{Peach}{\textbf{candidate generation}} module samples viewpoints on an uncertainty-adaptive ellipse around the predicted object-position mean.
    The \textcolor{MidnightBlue}{\textbf{reachability filtering}} module keeps only candidates reachable within one planning step.
    The \textcolor{OliveGreen}{\textbf{viewpoint evaluation}} module estimates each reachable candidate's expected coverage-driven score under the predictive belief.
    The selected action is the reachable candidate with the highest expected score; after execution, the new position measurement and point cloud are incorporated, and the loop repeats.}
    \label{fig:overview}
    \vspace{-0.15 in}
\end{figure}  

\section{Problem Setup} 
\label{sec:setup}

We consider active reconstruction of an unknown rigid object undergoing planar motion.
At each discrete planning time $t_k$, the planner receives two inputs: a depth observation from the sensing robot's calibrated camera and a noisy planar object-position measurement from a separate sensing source.  

\subsection{Notation and Reference Frames}

Let $\mathbf{x}^o(t)$ denote the continuous-time object state, with $\mathbf{x}^o_k \coloneqq \mathbf{x}^o(t_k)$ its value at planning time $t_k$.
Measurement incorporation, estimation updates, viewpoint selection, and execution are indexed by the discrete planning times $\{t_k\}$.

The world, robot base, and camera optical frames are denoted by $\mathcal{W}$, $\mathcal{R}$, and $\mathcal{C}$, respectively.
The reconstruction proxy (Sec.~\ref{sec:pb_nbv}) is maintained in an object-centric map frame $\mathcal{O}_k$ at $t_k$.
During planning at time $t_k$, a temporary predicted object frame $\tilde{\mathcal{O}}_{k+1}$ is created for the next planning time $t_{k+1}$.
This frame is derived from the one-step predictive belief and is used only to select viewpoints for execution at $t_{k+1}$.

After the robot executes the selected motion and the planner incorporates the new object-position measurement at $t_{k+1}$, the next object-centric map frame $\mathcal{O}_{k+1}$ is defined and the voxel map is updated in that frame.
Because $\tilde{\mathcal{O}}_{k+1}$ is defined from the prediction before the measurement update, whereas $\mathcal{O}_{k+1}$ is defined after incorporating the measurement at $t_{k+1}$, the two frames generally do not coincide.

The rigid-body transform that maps coordinates expressed in frame $\mathcal{B}$ into frame $\mathcal{A}$ is denoted by ${}^{\mathcal{A}}\mathbf{T}_{\mathcal{B}} \in SE(3)$, with rotation ${}^{\mathcal{A}}\mathbf{R}_{\mathcal{B}}\in SO(3)$ and translation ${}^{\mathcal{A}}\mathbf{t}_{\mathcal{B}}\in\mathbb{R}^3$.
 
\subsection{Robot model}
At planning time $t_k$, the robot's planar pose is denoted by
\begin{equation} 
    \mathbf x^r_k \coloneqq [\mathbf p^r_k,\;\theta^r_k]^\top \in \mathbb R^2\times S^1.
\end{equation}
The camera pose is obtained from the robot base pose ${}^{\mathcal W}\mathbf{T}_{\mathcal R}$ and known extrinsics ${}^{\mathcal R}\mathbf{T}_{\mathcal C}$ as $^{\mathcal W}\mathbf{T}_{\mathcal C} = {}^{\mathcal W}\mathbf{T}_{\mathcal R}{}^{\mathcal R}\mathbf{T}_{\mathcal C}$.

\subsection{Object state and measurements} 
The object state consists of planar position and velocity,
\begin{equation} 
    \mathbf x^o(t) \coloneqq [
    \mathbf p^o(t), \; 
    \dot{\mathbf p}^o(t)]^\top \in\mathbb R^4.
    \label{eq:object_state}
\end{equation}
At each planning time $t_k$, a separate sensing source provides a noisy measurement of the object's planar position,
\begin{equation} 
    \tilde{\mathbf p}^o_k = \mathbf p^o_k + \mathbf n_k,
    \qquad
    \mathbf n_k\sim\mathcal N(\mathbf 0,\mathbf R),
\end{equation}  
where $\mathbf R=\sigma^2\mathbf I \in\mathbb R^{2\times2}$ is the planar position measurement-noise covariance.
Velocity is not directly observed and is inferred as a latent component of the state via fixed-lag GP smoothing (Sec.~\ref{sec:pose_prediction}).
The object heading is assumed to align with the object's tangential direction of motion.


\section{Methodology} 

We use an online receding-horizon planning loop that combines object-state prediction with coverage-driven NBV selection (Fig.~\ref{fig:overview}).
At each planning time $t_k$, the planner incorporates the latest depth and object-position measurements, propagates the object state one step ahead to $t_{k+1}=t_k+\Delta t$ (Sec.~\ref{sec:pose_prediction}), and samples candidate viewpoints around the predicted object location and filters them by reachability (Sec.~\ref{sec:cand_generation}). 
Each reachable viewpoint is then evaluated by Monte Carlo estimation of its PB-NBV score under the predictive object-state belief (Sec.~\ref{sec:pb_nbv} and Sec.~\ref{sec:nbv_view_selection}).
The viewpoint with the highest expected score is executed, and the loop repeats as new measurements arrive.

\subsection{Object State Estimation and Prediction}
\label{sec:pose_prediction}
At each planning time $t_k$, a fixed-lag sparse GP smoother estimates a Gaussian belief over the object state $\mathbf{x}^o_k$ and predicts this belief one step ahead to $t_{k+1}$.
Because only planar position is measured, velocity is treated as a latent state.
The resulting predictive belief is used for candidate viewpoint generation and evaluation by the online NBV planner.
 
\subsubsection{Fixed-lag GP-based smoother}
Following sparse GP trajectory estimation~\cite{Barfoot13,Barfoot14}, the object trajectory is modeled using a white-noise acceleration constant-velocity prior,
\begin{equation} \label{eq:acc_prior}
    \ddot{\mathbf p}^o(t)=\mathbf w(t),\qquad
    \mathbf w(t)\sim\mathcal{GP}\!\big(\mathbf 0,\mathbf Q_C\,\delta(t-t')\big), 
\end{equation} 
where $\mathbf Q_C=q_c\mathbf I\in\mathbb R^{2\times2}$ is the continuous-time acceleration-noise power spectral density, and $\delta(\cdot)$ denotes the Dirac delta.

At time $t_k$, the position-only measurement model is
\begin{equation*}
    p(\tilde{\mathbf p}^o_k \mid \mathbf x^o_k) = \mathcal N(\tilde{\mathbf p}^o_k;\, \mathbf H\mathbf x^o_k,\mathbf R),
    \qquad
    \mathbf H=[\mathbf I\ \mathbf 0]\in\mathbb R^{2\times 4}.
\end{equation*}  
A fixed-lag Maximum A Posteriori smoother is then applied to the most recent $L$ measurements $\mathcal D_k=\{\tilde{\mathbf p}^o_{k-L+1},\ldots,\tilde{\mathbf p}^o_k\}$. 

Let $\boldsymbol\mu^o_k$ and $\boldsymbol\Sigma^o_k$ denote the marginal mean and covariance of the current object state $\mathbf x^o_k$ obtained from the fixed-lag posterior; the current belief is
\begin{equation}\label{eq:belief_current} 
    p(\mathbf x^o_k\mid \mathcal D_k)=\mathcal N\big(\mathbf x^o_k;\,\boldsymbol\mu^o_k,\boldsymbol\Sigma^o_k\big).
\end{equation}

\subsubsection{One-step GP predictor}
For a time increment $\Delta t$, the motion prior in Eq.~\ref{eq:acc_prior} yields the closed-form transition
\begin{equation*}
    \boldsymbol\Phi(\Delta t)
    =
    \begin{bmatrix}
    \mathbf I & \Delta t \mathbf I\\
    \mathbf 0 & \mathbf I
    \end{bmatrix},
    \quad\;\;\;
    \mathbf Q(\Delta t)=
    \begin{bmatrix} 
        \frac{1}{3}\Delta t^3 \mathbf Q_C & \frac{1}{2}\Delta t^2 \mathbf Q_C\\
        \frac{1}{2}\Delta t^2 \mathbf Q_C & \Delta t \mathbf Q_C
    \end{bmatrix}.
\end{equation*} 

Given the current belief $p(\mathbf x^o_k\mid \mathcal D_k)$ from Eq.~\ref{eq:belief_current}, the one-step predictive mean and covariance are
\begin{align}
\boldsymbol\mu^o_{k+1|k} &= \boldsymbol\Phi(\Delta t)\,\boldsymbol\mu^o_k,
\label{eq:mean_prediction}\\
\boldsymbol\Sigma^o_{k+1|k} &= \boldsymbol\Phi(\Delta t)\,\boldsymbol\Sigma^o_k\,\boldsymbol\Phi(\Delta t)^\top + \mathbf Q(\Delta t).
\label{eq:cov_prediction}
\end{align}
Candidate viewpoint generation and evaluation at planning step $k$ are conditioned on the predictive Gaussian
\begin{equation}
    p(\mathbf x^o_{k+1}\mid \mathcal D_k)
    =
    \mathcal N\big(\mathbf x^o_{k+1};\,\boldsymbol\mu^o_{k+1|k},\boldsymbol\Sigma^o_{k+1|k}\big).
    \label{eq:predictive_belief}
\end{equation}
This predictive belief contains both planar position and velocity components, together with their covariance.

\subsubsection{Object nominal heading}
Following the tangential-heading assumption in Sec.~\ref{sec:setup}, the object's nominal heading is derived from the predicted mean velocity:
\begin{equation}\label{eq:nominal_heading}
    \hat{\theta}^o_{k+1} = \operatorname{atan2}\!\big((\boldsymbol\mu^o_{\dot{\mathbf p},\,k+1|k})_y,(\boldsymbol\mu^o_{\dot{\mathbf p},\,k+1|k})_x\big),
\end{equation}
where $\boldsymbol\mu^o_{\dot{\mathbf p},\,k+1|k}$ denotes the velocity block of the predicted mean in~\eqref{eq:mean_prediction}.
This expression assumes $\|\boldsymbol\mu^o_{\dot{\mathbf p},\,k+1|k}\|>0$; when the speed is below a small threshold, the previous nominal heading is retained.
The nominal heading $\hat{\theta}^o_{k+1}$ is used only to define the predicted scoring frame $\widetilde{\mathcal O}_{k+1}$ for candidate generation (Sec.~\ref{sec:cand_generation});
uncertainty in the object state is handled explicitly during viewpoint evaluation (Sec.~\ref{sec:nbv_view_selection}).


\subsection{Prediction-Conditioned Candidate Viewpoint Generation}
\label{sec:cand_generation}
Using the predictive belief in Eq.~\ref{eq:predictive_belief}, we sample candidate viewpoints on an uncertainty-adaptive ellipse that preserves a desired stand-off distance while expanding along directions of larger positional uncertainty.

\subsubsection{Uncertainty-adaptive ellipse discretization}
\label{sec:ellipse_discretization}

We define a predicted object-centric scoring frame $\widetilde{\mathcal O}_{k+1}$ for viewpoint generation and evaluation at the future time $t_{k+1}$.
Let
\begin{equation}  
    \mathbf c_{k+1} \coloneqq \boldsymbol\mu_{\mathbf p,\,k+1|k}^o
    \label{eq:pred_mean}
\end{equation}
denote the predicted mean position extracted from the position block of~\eqref{eq:mean_prediction}.
The planar origin of $\widetilde{\mathcal O}_{k+1}$ is located at $\mathbf c_{k+1}$, and its $x$-axis is aligned with the nominal predicted heading derived in Eq.~\ref{eq:nominal_heading}. 
This defines the planar rotation
\begin{equation*}  
{}^{\mathcal W}\!\mathbf R_{\widetilde{\mathcal O},k+1}\coloneqq
\begin{bmatrix}
\cos(\hat{\theta}^o_{k+1}) & -\sin(\hat{\theta}^o_{k+1})\\
\sin(\hat{\theta}^o_{k+1}) & \cos(\hat{\theta}^o_{k+1})
\end{bmatrix}\in SO(2).
\end{equation*}

\begin{figure*}[t]
  \centering
  \includegraphics[width =0.95\linewidth, trim=0 0 0 0]{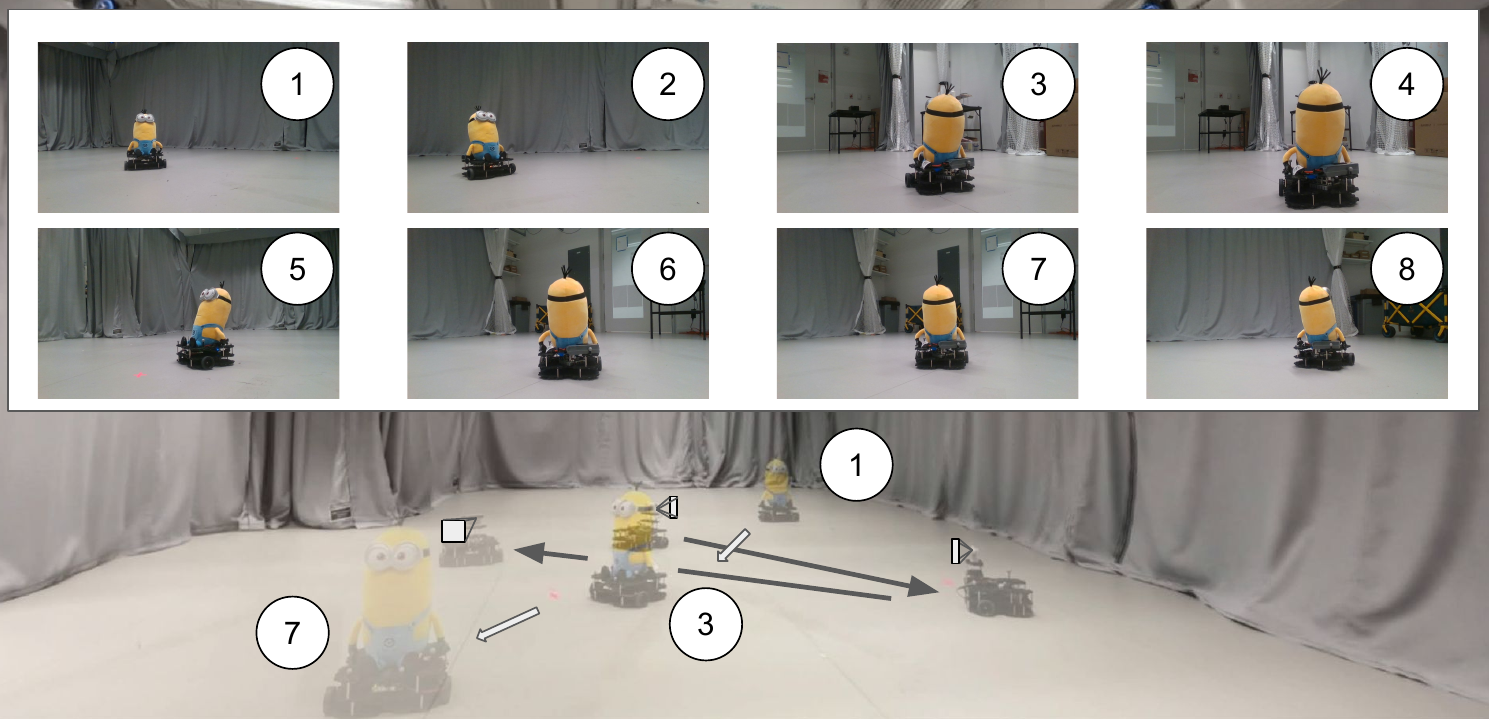}
  \caption{\textbf{Real-world replanning sequence from one representative trial.} 
  \textbf{Top}: RGB-camera views acquired over eight successive planning iterations for visual illustration.  
  \textbf{Bottom}: corresponding robot and object configurations overlaid in a common frame; numbered markers indicate selected iterations. The sequence shows that the robot does not simply follow the moving object, but changes perspectives to observe new object surfaces as reconstruction progresses.}
  \label{fig:real_experiment}
  \vspace{-0.20 in}
\end{figure*} 

Inspired by uncertainty-aware observation strategies~\cite{liu25}, we generate candidate viewpoints on a planar ellipse centered at $\mathbf c_{k+1}$~\eqref{eq:pred_mean}, with semi-axes adapted to the predicted positional uncertainty.
The idea is to enlarge the candidate search radius along directions where the future object position is more uncertain, while preserving a nominal stand-off distance when uncertainty is low.
Let $\boldsymbol\Sigma_{\mathbf p,\,k+1|k}^o\in\mathbb R^{2\times2}$ denote the position block of the predicted covariance in~\eqref{eq:cov_prediction}. 
The positional covariance expressed in the coordinates of $\widetilde{\mathcal O}_{k+1}$ is
\begin{equation*} 
    \tilde{\boldsymbol\Sigma}_{\mathbf p,\,k+1|k}^o
    \coloneqq
    ({}^{\mathcal W}\!\mathbf R_{\widetilde{\mathcal O},k+1})^\top
    \boldsymbol\Sigma_{\mathbf p,\,k+1|k}^o
    {}^{\mathcal W}\!\mathbf R_{\widetilde{\mathcal O},k+1}.
    \label{eq:cov_rotate_heading}
\end{equation*}

We use the diagonal of $\tilde{\boldsymbol\Sigma}_{\mathbf p,\,k+1|k}^o$ as heading-aligned marginal variances and define forward and lateral standard deviations as the square roots of the diagonal entries,
\begin{equation*} \label{eq:sigma_parallel_perp}
    \sigma_{\parallel,k+1}\coloneqq \sqrt{\big(\tilde{\boldsymbol\Sigma}_{\mathbf p,\,k+1|k}^o\big)_{11}},
    \qquad
    \sigma_{\perp,k+1}\coloneqq \sqrt{\big(\tilde{\boldsymbol\Sigma}_{\mathbf p,\,k+1|k}^o\big)_{22}}.
\end{equation*}
The ellipse semi-axes are given by 
\begin{equation*} \label{eq:ellipse_axes}
    a_{k+1}\coloneqq r_{\mathrm{off}}+\kappa\,\sigma_{\parallel,k+1},
    \qquad
    b_{k+1}\coloneqq r_{\mathrm{off}}+\kappa\,\sigma_{\perp,k+1},
\end{equation*}
where $r_{\mathrm{off}}>0$ is a desired stand-off distance chosen from the camera sensing range, and $\kappa>0$ is an uncertainty-inflation factor that trades off coverage of likely object locations against reachability.
Uncertainty is reflected through anisotropic scaling of the viewpoint generation ellipse axes, while the ellipse orientation is governed by the nominal motion direction (Eq.~\ref{eq:nominal_heading}) for stable and interpretable viewpoint placement.

The ellipse is discretized using $I$ uniformly spaced azimuth angles $\varphi_i = 2\pi(i-1)/I$, for $i\in\{1,\ldots,I\}$.
Each index $i$ defines one candidate viewpoint on the ellipse.

\subsubsection{Candidate viewpoint positions and yaw}
\label{sec:cand_viewpoint}

For each azimuth index $i$, the planar offset in the predicted object-centric scoring frame $\widetilde{\mathcal O}_{k+1}$ is defined as
\begin{equation*} \label{eq:ellipse_offset}
    \mathbf q_{k+1,i}
    \coloneqq 
    \begin{bmatrix}
    a_{k+1}\cos\varphi_i\\
    b_{k+1}\sin\varphi_i
    \end{bmatrix}
    \in\mathbb R^2.
\end{equation*}
The corresponding planar candidate position in $\mathcal W$ is then
\begin{equation} 
    \mathbf p^{\mathrm{cand}}_{k+1,i}
    \coloneqq
    \mathbf c_{k+1} + {}^{\mathcal W}\!\mathbf R_{\widetilde{\mathcal O},k+1}\mathbf q_{k+1,i}
    \in\mathbb R^2 ,
    \label{eq:cand_position}
\end{equation}
with an assigned inward-facing yaw that points toward the predicted object center $\mathbf c_{k+1}$~\eqref{eq:pred_mean}:
\begin{equation} \label{eq:cand_yaw}
\psi^{\mathrm{cand}}_{k+1,i}
\coloneqq
\operatorname{atan2}\!\Big(
(\mathbf c_{k+1}-\mathbf p^{\mathrm{cand}}_{k+1,i})_y,\;
(\mathbf c_{k+1}-\mathbf p^{\mathrm{cand}}_{k+1,i})_x
\Big).
\end{equation}

Embedding the planar candidate position and yaw as a robot-base pose with zero roll and pitch gives
\begin{equation*} 
    {}^{\mathcal W}\mathbf T^{\mathrm{cand}}_{\mathcal R,k+1,i}
    \coloneqq
    \begin{pmatrix}
    R_z\!\big(\psi^{\mathrm{cand}}_{k+1,i}\big) &
    \begin{bmatrix}
    \mathbf p^{\mathrm{cand}}_{k+1,i}\\
    0
    \end{bmatrix}\\
    \mathbf 0^\top & 1
    \end{pmatrix}.
\end{equation*}
Applying the known camera extrinsics ${}^{\mathcal R}\mathbf T_{\mathcal C}$ yields the corresponding candidate camera pose
\begin{equation*} 
    {}^{\mathcal W}\mathbf T^{\mathrm{cand}}_{\mathcal C,k+1,i}
    \coloneqq
    {}^{\mathcal W}\mathbf T^{\mathrm{cand}}_{\mathcal R,k+1,i}
    {}^{\mathcal R}\mathbf T_{\mathcal C}
    \in SE(3).
\end{equation*} 
Collecting all camera poses, we obtain the candidate set
\begin{equation*} \label{eq:cand_set}
    \mathbb T^{\mathrm{cand}}_{k+1}
    \coloneqq
    \left\{
    {}^{\mathcal W}\mathbf T^{\mathrm{cand}}_{\mathcal C,k+1,i}
    \;\middle|\;
    i\in\{1,\ldots,I\}
    \right\}.
\end{equation*}

\subsubsection{Reachability filtering}
\label{sec:view_selection} 
Not all candidate viewpoints are attainable within a single planning step. We therefore filter candidates according to the current robot state $\mathbf x^r_k$ and bounded unicycle motion limits.
A candidate with planar position $\mathbf p^{\mathrm{cand}}_{k+1,i}$~\eqref{eq:cand_position} and yaw $\psi^{\mathrm{cand}}_{k+1,i}$~\eqref{eq:cand_yaw} is feasible if 
\begin{equation}
\|\mathbf p^{\mathrm{cand}}_{k+1,i}-\mathbf p^r_k\|
\le v_{\max}\,\Delta t,
\quad
|\psi^{\mathrm{cand}}_{k+1,i}-\theta^r_k|
\le \omega_{\max}\,\Delta t.
\label{eq:reachability}
\end{equation}
The candidates satisfying~\eqref{eq:reachability} form the feasible set $\mathbb T^{\mathrm{feas}}_{k+1}\subseteq\mathbb T^{\mathrm{cand}}_{k+1}$, which is passed to the NBV evaluation in Sec.~\ref{sec:nbv_view_selection}.


\subsection{Coverage-Driven NBV Evaluation}
\label{sec:pb_nbv}
The objective is to improve surface coverage of an initially unknown object during online replanning.
We therefore use an efficient coverage-driven NBV proxy that scores viewpoints by the amount of new surface they are expected to reveal from an object-centric bounded volumetric representation.

\subsubsection{Object-centric bounded voxel proxy}
As the robot moves, point clouds are registered to $\mathcal O_k$ and incrementally fused into a bounded object-centric voxel grid.
This bounded-domain representation is widely used in active 3D reconstruction to keep mapping and viewpoint evaluation tractable~\cite{Jia25,Delmerico18,Lauri20}.
Within this voxel proxy, voxels are labeled as \textit{Occupied} (observed surface), \textit{Free} (empty space), and \textit{Unknown} (unobserved). 
\textit{Frontier} voxels are extracted at the boundary between the observed region, consisting of Occupied and Free voxels, and the Unknown region, indicating where additional measurements are likely to add new information~\cite{Hepp17,Duberg20,Dai20}.

\subsubsection{Efficient voxel map proxy}
Given a fixed relative pose ${}^{\mathcal O}\mathbf T_{\mathcal C}$, we evaluate a deterministic coverage-driven viewpoint score $S({}^{\mathcal O}\mathbf T_{\mathcal C})$.
In our implementation, $S(\cdot)$ is instantiated using the projection-based PB-NBV score~\cite{Jia25} computed from the current object-centric voxel proxy.
Our framework is agnostic to the choice of deterministic coverage-driven NBV score; PB-NBV is used as an efficient per-view scoring surrogate.

We consider two voxel sets from the bounded voxel grid: Frontier voxels and Occupied voxels.
Following PB-NBV~\cite{Jia25}, each voxel set is clustered using Gaussian mixture models, and each cluster is approximated by a minimum-volume enclosing ellipsoid.
This yields two ellipsoid collections, 
\begin{equation*}
    {}^{\mathcal O}\mathcal E_F=\{{}^{\mathcal O}E_{F,u}\}_{u=1}^{K_F}, \qquad {}^{\mathcal O}\mathcal E_O=\{{}^{\mathcal O}E_{O,v}\}_{v=1}^{K_O}
\end{equation*}
where $K_F$ and $K_O$ are the numbers of Frontier and Occupied clusters, respectively.
These collections summarize where additional surface is likely to be revealed and where surface has already been observed.
 
\subsubsection{Projection-based pixel-area score}
For a candidate relative pose ${}^{\mathcal O}\mathbf T_{\mathcal C}$, each ellipsoid is transformed to the camera frame using
${}^{\mathcal C}\mathbf T_{\mathcal O}=({}^{\mathcal O}\mathbf T_{\mathcal C})^{-1}$.
A point $\mathbf x_{\mathcal O}$ on ellipsoid $E$ is mapped to the camera frame as
$\mathbf x_{\mathcal C}={}^{\mathcal C}\mathbf R_{\mathcal O}\mathbf x_{\mathcal O}+{}^{\mathcal C}\mathbf t_{\mathcal O}$
and projected into the image using the calibrated intrinsics $\mathbf K$:
\begin{equation} \label{eq:projection}
    \tilde{\mathbf u}\sim \mathbf K\Big({}^{\mathcal C}\mathbf R_{\mathcal O}\,\mathbf x_{\mathcal O} + {}^{\mathcal C}\mathbf t_{\mathcal O}\Big),
    \quad
    (\mathbf x_{\mathcal O}-\mathbf m_E)^\top \mathbf A_E
    (\mathbf x_{\mathcal O}-\mathbf m_E) \le 1,
\end{equation}
where $\tilde{\mathbf u}=(u,v,1)^\top$ denotes homogeneous image coordinates, and $\mathbf m_E$ and $\mathbf A_E$ denote the ellipsoid center and shape matrix in $\mathcal O$, respectively.
The projected silhouette region $\Omega_E$ is obtained by projecting the ellipsoid into the image and rasterizing the filled image region~\cite{besl92}.
Large projected Frontier regions correspond to viewpoints expected to observe more currently unknown surface, whereas large projected Occupied regions indicate views dominated by already observed surface.

To approximate visibility ordering along the camera optical axis, the ellipsoids
${}^{\mathcal O}\mathcal E_F \cup {}^{\mathcal O}\mathcal E_O$
are sorted by increasing depth.
Let $r(E)$ denote the resulting depth rank of ellipsoid $E$, with $r(E)=1$ corresponding to the closest ellipsoid.
A depth-ordering weight is assigned as
\begin{equation}\label{eq:depth_order}
    w_E=\alpha^{\,r(E)-1},\qquad \alpha\in(0,1],
\end{equation}
so that farther ellipsoids contribute less to the score.
Let $\Omega_{F,u}$ and $\Omega_{O,v}$ denote the projected image regions of Frontier and Occupied ellipsoids, with corresponding weights $w_{F,u}$ and $w_{O,v}$. 
The deterministic PB-NBV score~\cite{Jia25} is then given by
\begin{equation}
    S\!\left({}^{\mathcal O}\mathbf T_{\mathcal C}\right) =
    \sum_{u=1}^{K_F} w_{F,u}\,|\Omega_{F,u}|-
    \sum_{v=1}^{K_O} w_{O,v}\,|\Omega_{O,v}|,
    \label{eq:det_score} 
\end{equation}
where $|\Omega|$ denotes the projected pixel area.
The score rewards projected Frontier area, which approximates newly observable surface, and penalizes projected Occupied area, which corresponds to already observed surface.


\subsection{Motion-Uncertainty-Aware NBV Planning}
\label{sec:nbv_view_selection}

We select the NBV under uncertainty in the object state at execution time $t_{k+1}$.
The key idea is to evaluate candidate viewpoints under the distribution of future camera-object configurations induced by the predictive object-state belief, rather than at a single predicted pose.

\subsubsection{Camera-object configuration under belief sampling}
We draw $N$ i.i.d.\ samples
$\{\mathbf x^{o,(n)}_{k+1}\}_{n=1}^N$
from the predictive belief in Eq.~\ref{eq:predictive_belief}.
From each sample, the planar position $\mathbf p^{o,(n)}_{k+1}$ and velocity $\dot{\mathbf p}^{o,(n)}_{k+1}$ are extracted. For samples with nonzero speed, the corresponding heading is
\begin{equation*}
    \theta^{o,(n)}_{k+1} = \operatorname{atan2}\!\Big((\dot{\mathbf p}^{o,(n)}_{k+1})_y,\;(\dot{\mathbf p}^{o,(n)}_{k+1})_x\Big).
\end{equation*}
When the sampled speed is below a small threshold, the nominal heading from Eq.~\ref{eq:nominal_heading} is retained.
Each sample then induces a world-frame pose of the predicted object frame,
\begin{equation*}
    {}^{\mathcal W}\mathbf T^{(n)}_{\widetilde{\mathcal O},k+1}
    \coloneqq
    \begin{pmatrix}
    R_z\!\big(\theta^{o,(n)}_{k+1}\big) &
    \begin{bmatrix}
    \mathbf p^{o,(n)}_{k+1}\\
    0
    \end{bmatrix}\\
    \mathbf 0^\top & 1
    \end{pmatrix}
    \in SE(3).
\label{eq:T_WO_sample}
\end{equation*}

For a fixed candidate camera pose
${}^{\mathcal W}\mathbf T^{\mathrm{cand}}_{\mathcal C,k+1}$,
the sample-induced relative transform is
\begin{equation*}
{}^{\widetilde{\mathcal O}^{(n)}_{k+1}}\mathbf T^{\mathrm{cand},(n)}_{\mathcal C,k+1}
    \coloneqq
    \big({}^{\mathcal W}\mathbf T^{(n)}_{\widetilde{\mathcal O},k+1}\big)^{-1}
    {}^{\mathcal W}\mathbf T^{\mathrm{cand}}_{\mathcal C,k+1}.
    \label{eq:relative_transform}
\end{equation*} 
With known robot pose, object-state uncertainty appears as variability in this relative camera-object configuration.


\begin{figure*}[ht]
  \centering
  \includegraphics[width=\linewidth, trim=40 30 0 30]{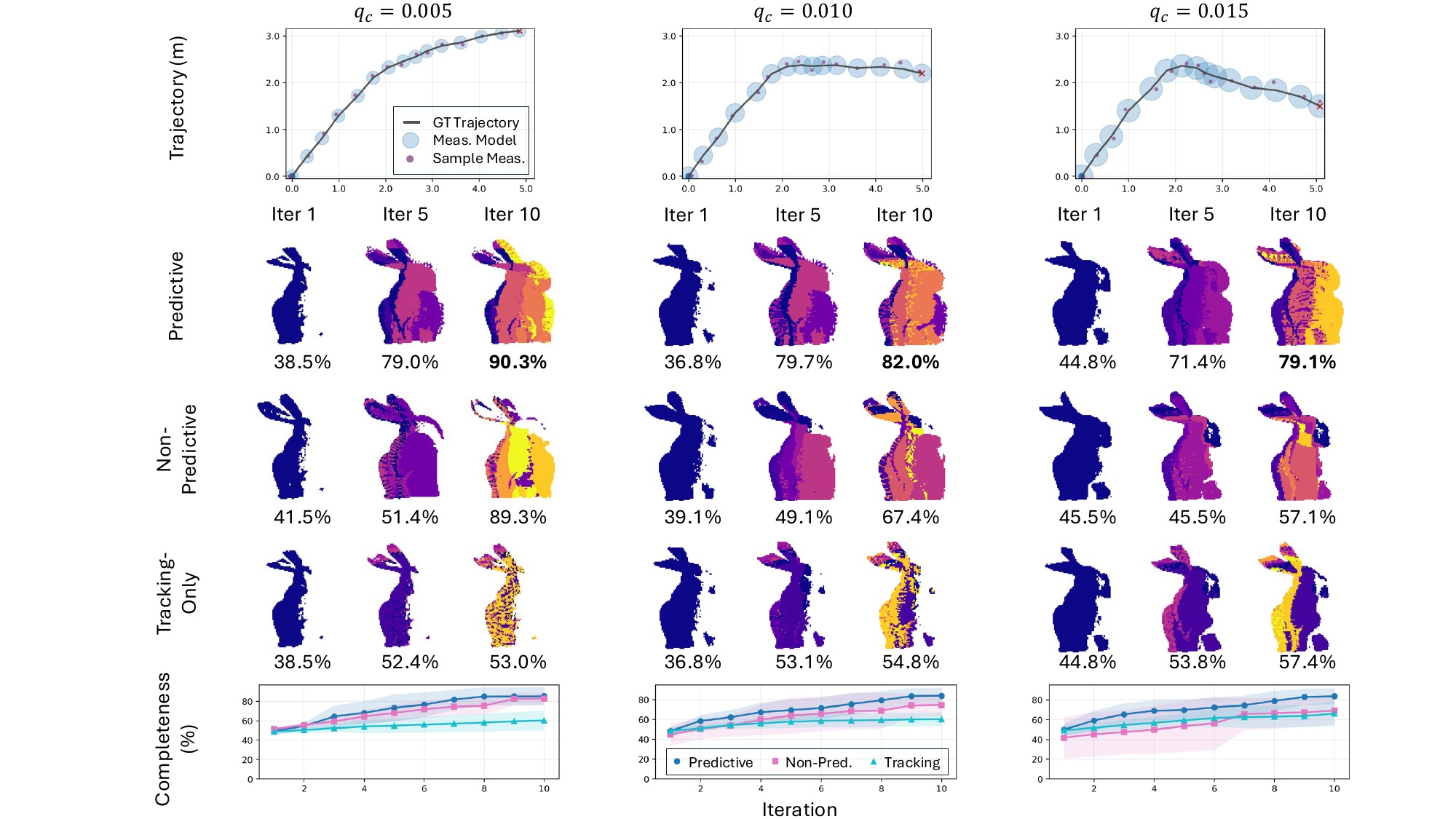}
  \caption{\textbf{Simulation reconstruction under varying process noise.}
  The process-noise level increases across columns, $q_c\in\{0.005,0.010,0.015\}$, with measurement noise fixed at $0.10$\,m.
  \textbf{Top:} representative ground-truth (GT) object trajectories, position-uncertainty regions, and noisy position measurements.
  \textbf{Middle:} representative reconstructions at iterations 1, 5, and 10 for the three viewpoint selection methods; numbers report reconstruction completeness, and colors indicate the surface revealed by each iteration.
  \textbf{Bottom:} completeness over planning iterations, reported as mean $\pm$ std over 10 random seeds.}
  \label{fig:sim_progress}
  \vspace{-0.15 in}
\end{figure*}


\begin{figure}[t]
    \centering
    \includegraphics[width = \linewidth, trim=10 260 15 -10]
    {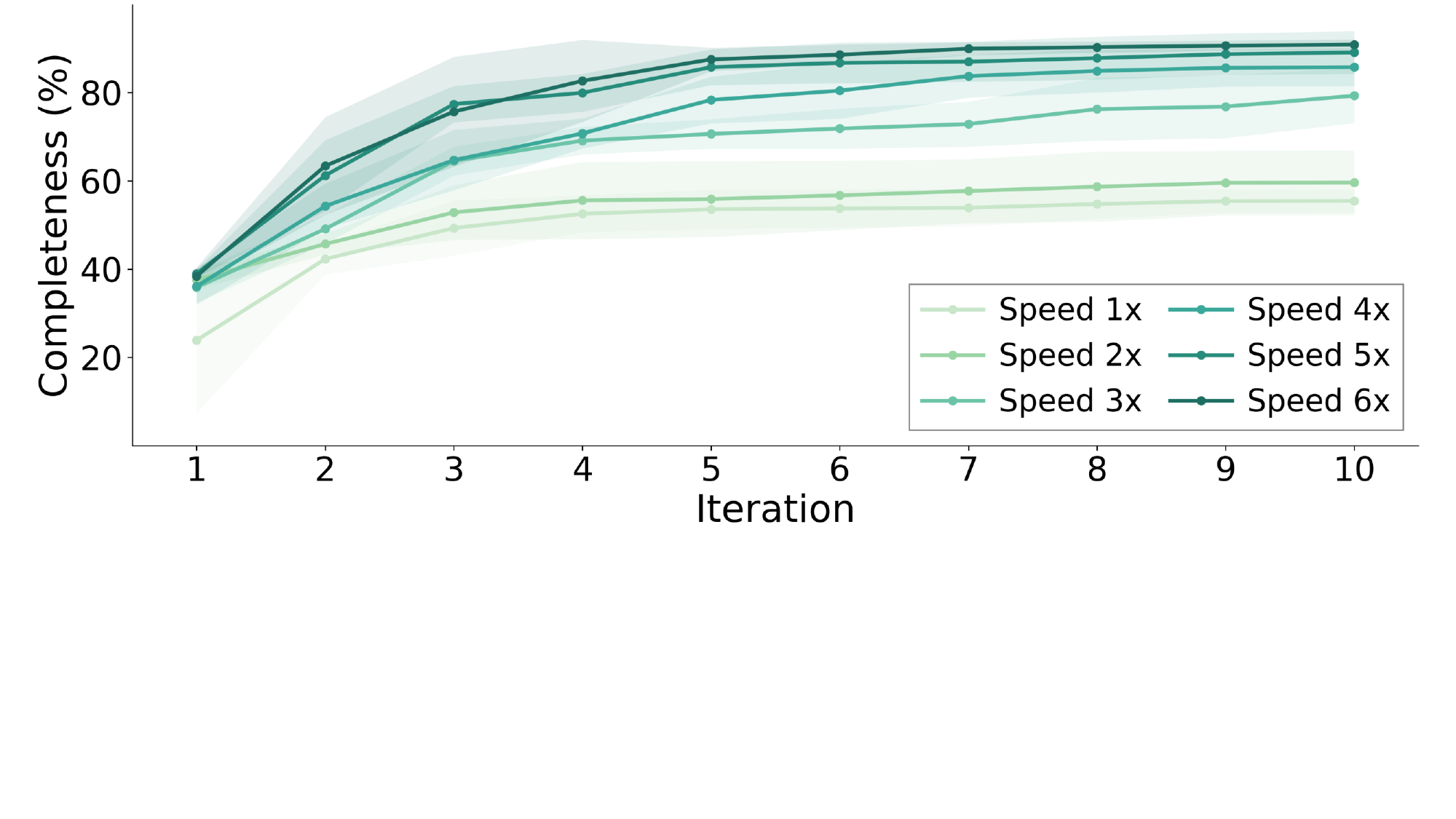}
    \caption{\textbf{Effect of robot-object relative speed in simulation.} 
    The speed factor scales the robot's maximum per-step travel distance relative to the object's displacement. Curves report reconstruction completeness as mean $\pm$ std over 10 random-seeded trajectories with fixed process and measurement noise.}
    \label{fig:coverage_vs_speed}
  \vspace{-0.20 in}
\end{figure} 


\subsubsection{Expected NBV score}
\label{sec:uncertainty_score}

For each feasible candidate
${}^{\mathcal W}\mathbf T^{\mathrm{cand}}_{\mathcal C,k+1}\in\mathbb T^{\mathrm{feas}}_{k+1}$,
the uncertainty-aware score is defined as the predictive expectation of the deterministic NBV score in Eq.~\ref{eq:det_score}.
For an object state $\mathbf x^o_{k+1}$, let
${}^{\mathcal W}\mathbf T_{\widetilde{\mathcal O}}(\mathbf x^o_{k+1})$
denote the predicted object-frame pose constructed from its position and velocity using the tangential-heading rule above.
For a fixed candidate pose, the induced relative transform is
\begin{equation*}
    {}^{\widetilde{\mathcal O}}\mathbf T_{\mathcal C}^{\mathrm{rel}}(\mathbf x^o_{k+1})
    \coloneqq
    \left({}^{\mathcal W}\mathbf T_{\widetilde{\mathcal O}}(\mathbf x^o_{k+1})\right)^{-1}
    {}^{\mathcal W}\mathbf T^{\mathrm{cand}}_{\mathcal C,k+1}.
\end{equation*}

The motion-uncertainty-aware NBV score is then
\begin{equation}
    J_{k+1}\!\left({}^{\mathcal W}\mathbf T^{\mathrm{cand}}_{\mathcal C,k+1}\right)
    \coloneqq
    \mathbb{E}_{\mathbf x^o_{k+1} }
    \!\left[
    S\!\left(
    {}^{\widetilde{\mathcal O}}\mathbf T_{\mathcal C}^{\mathrm{rel}}(\mathbf x^o_{k+1})
    \right)
    \right].
    \label{eq:score_integral}
\end{equation}
This expectation admits no closed form because $S(\cdot)$ (Eq.~\ref{eq:det_score}) depends on nonlinear camera projection, discretization, and depth-ordering operations
(Eq.~\ref{eq:projection} and Eq.~\ref{eq:depth_order}), which make the integrand in Eq.~\ref{eq:score_integral} non-smooth in the object state.
By the law of large numbers, belief-space samples from $p(\mathbf x^o_{k+1}\mid\mathcal D_k)$ provide a consistent Monte Carlo estimate of the predictive expectation.
We therefore approximate Eq.~\ref{eq:score_integral} by Monte Carlo sampling:
\begin{equation} 
    J_{k+1}\!\left({}^{\mathcal W}\mathbf T^{\mathrm{cand}}_{\mathcal C,k+1}\right)
    \approx
    \frac{1}{N}\sum_{n=1}^{N}
    S\!\left(
    {}^{\widetilde{\mathcal O}}\mathbf T_{\mathcal C}^{\mathrm{rel}}(\mathbf x^{o,(n)}_{k+1})
    \right).
    \label{eq:mc_score}
\end{equation}
 
\subsubsection{Motion-Uncertainty-Aware NBV Selection}
\label{sec:uncertainty_score_selection}
The next viewpoint is selected by maximizing the expected motion-uncertainty-aware NBV score over the feasible candidate set:
\begin{equation}
    {}^{\mathcal W}\mathbf T^{\star}_{\mathcal C,k+1}
    \coloneqq
    \argmax_{
    {}^{\mathcal W}\mathbf T^{\mathrm{cand}}_{\mathcal C,k+1}\in
    \mathbb T^{\mathrm{feas}}_{k+1}
    }
    J_{k+1}\!\left({}^{\mathcal W}\mathbf T^{\mathrm{cand}}_{\mathcal C,k+1}\right).
    \label{eq:greedy_select}
\end{equation}
This policy favors viewpoints with high expected coverage-driven score under future object-state uncertainty. 

\begin{tcolorbox} [
      colback=gray!12,
      colframe=black!70,
      boxrule=1pt,
      arc=2mm,
      left=2mm,
      right=2mm,
      top=1mm,
      bottom=1mm
    ]
    {\centering\bfseries Remark\par}
    \vspace{1mm}
    The method uses object-state prediction twice: (i) to shape candidate geometry according to the object-state uncertainty, and (ii) to evaluate viewpoints by expected observation quality over the predictive belief.
\end{tcolorbox}
 
\section{Experiments}

\subsection{Experimental Goals and Metrics}
The experiments evaluate whether motion-uncertainty-aware NBV improves coverage-driven reconstruction of a moving object. They are organized around three questions:
(i) whether one-step predictive belief-space evaluation improves reconstruction over scoring viewpoints at the current estimated object state;
(ii) whether coverage-driven viewpoint selection provides benefits beyond prediction-only tracking; and
(iii) how uncertainty-aware candidate generation and Monte Carlo belief-space evaluation contribute to performance.

We use a voxelized surface-completeness metric inspired by standard reconstruction completeness measures~\cite{Petrovska24}. 
For both simulation and real-world experiments, the ground-truth object point cloud is constructed by merging depth observations from robot-accessible viewpoints uniformly spaced around the object at $15^\circ$ angular increments.
At each planning iteration $k$, the accumulated reconstructed points are transformed into the object-centric evaluation frame and voxelized to form $V_{1:k}$.
The ground-truth object point cloud is voxelized at the same resolution to form $V_{\mathrm{gt}}$.
Completeness is defined as
\begin{equation*}
    \mathrm{Comp}(k)=100\cdot \frac{|V_{1:k}\cap V_{\mathrm{gt}}|}{|V_{\mathrm{gt}}|}.
\end{equation*}
This metric reports the percentage of ground-truth object voxels observed by the reconstruction.

Unless otherwise stated, all compared methods use the same object-position measurements, fixed-lag GP smoother window size, candidate discretization, reachability constraints, and voxel-map update procedure.
They differ only in how candidate viewpoints are generated, scored, or selected.

We compare the proposed method with two baselines representing the paradigms our framework bridges:
\begin{enumerate}
    \item \textbf{Predictive NBV (ours)} selects viewpoints by maximizing the expected coverage-driven score under the one-step predictive object-state belief.
    \item \textbf{Non-predictive NBV} evaluates viewpoints under the current object-state belief, without propagating the belief to execution time.
    \item \textbf{Tracking-only} uses the predicted object state but selects the feasible candidate closest to the predicted object location, without coverage-driven NBV scoring.
\end{enumerate}

The comparison with Non-predictive NBV isolates the effect of accounting for object motion during the decision-to-execution interval, while the comparison with Tracking-only isolates the effect of coverage-driven viewpoint selection beyond following the predicted target.

\subsection{Simulation Experiments}
\label{sec:sim}

The simulations evaluate how reconstruction performance changes with object process noise, robot reachability limits, and the main components of the proposed planner.
Experiments are run online in Gazebo on Ubuntu 24.04 with ROS 2 Jazzy, using a workstation equipped with a 12th Gen Intel Core i9-12900K.
All planning is performed on CPU, with subsecond computation per planning iteration.
A mobile robot with a depth camera replans viewpoints to reconstruct a rigid object undergoing planar motion.


\begin{figure}[t]
    \centering
    \includegraphics[width = \linewidth, trim=10 240 0 0]
    {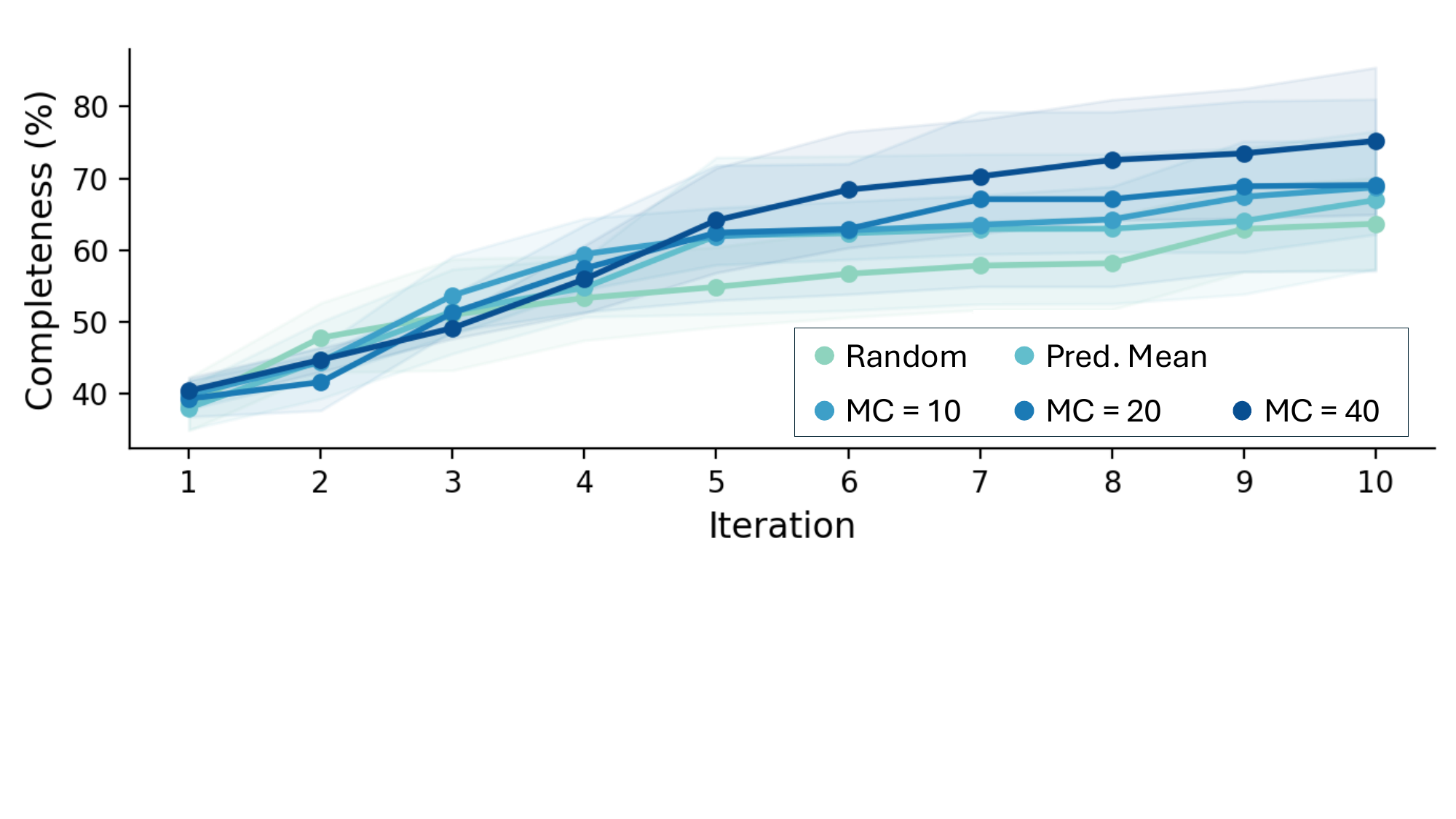}
    \caption{\textbf{Ablation of viewpoint evaluation under the predictive belief.}
    \emph{Random} selects uniformly from the feasible candidate set, \emph{Pred. Mean} scores only at the predicted object-state mean, and MC uses $N\in\{10,20,40\}$ samples from the one-step predictive belief.
    Curves report mean $\pm$ std over 10 seeds using the same trajectories and measurement realizations.}
  \label{fig:mc_coverage} 
  \vspace{-0.15 in}
\end{figure} 


\begin{figure*}[ht]
  \centering
  \includegraphics[width = \linewidth, trim=40 33 0 30]{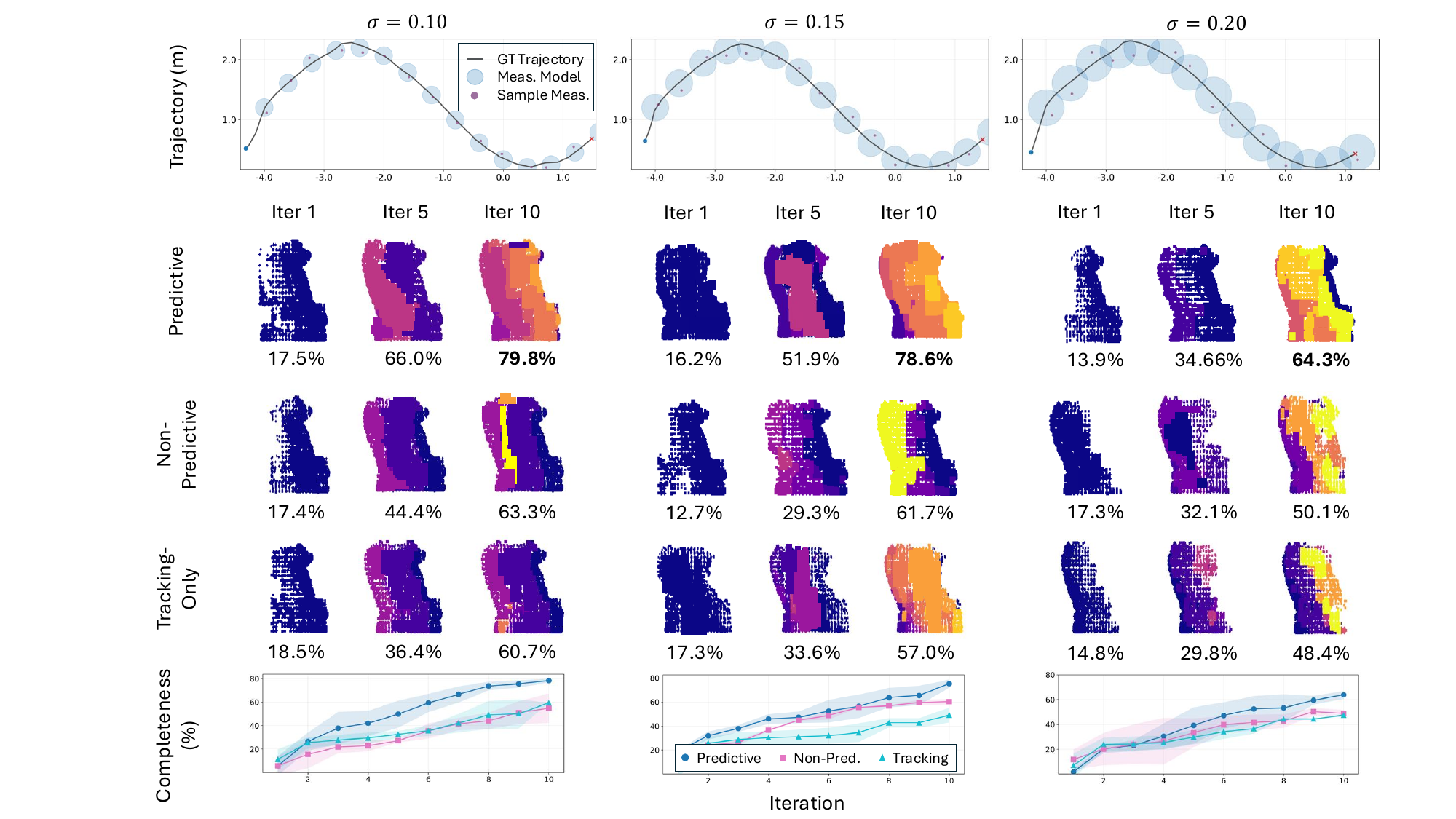}
  \caption{\textbf{Real-world reconstruction under varying measurement noise.}
  The injected measurement noise increases across columns, $\sigma\in\{0.10,0.15,0.20\}$\,m.
  \textbf{Top:} measured ground-truth (GT) object trajectory, position-uncertainty regions, and noisy position measurements.
  \textbf{Middle:} representative reconstructions at iterations 1, 5, and 10 for the three viewpoint selection methods; numbers report reconstruction completeness, and colors indicate the surface revealed by each iteration.
  \textbf{Bottom:} completeness over planning iterations, reported as mean $\pm$ std over four trials per setting.}
  \label{fig:real_progress}
  \vspace{-0.20 in}
\end{figure*} 


\textbf{Baseline comparison under process noise.}
Figure~\ref{fig:sim_progress} compares the three viewpoint selection methods as object process noise increases.
Our proposed Predictive NBV achieves the highest final completeness across all process-noise settings, with the largest gains at higher process noise.
The representative reconstructions explain this trend: Predictive NBV continues to reveal new surface regions in later iterations, while Tracking-only often stays near the predicted target without improving surface coverage.
The improvement over Non-predictive NBV shows that evaluating viewpoints under the current belief becomes unreliable when object motion changes the execution-time camera-object configuration.
The improvement over Tracking-only shows that prediction must be coupled with coverage-driven scoring to reveal previously unseen surfaces.
Overall, propagating the belief to execution time produces viewpoint choices that remain reconstruction-informative under plausible future object motion.

\textbf{Effect of robot reachability.} Figure~\ref{fig:coverage_vs_speed} evaluates how reachability limits affect reconstruction by varying the sensing robot's maximum per-step travel distance relative to the object's displacement, which is approximately $0.5\,\mathrm{m}$ per planning step.
Higher relative speed improves both coverage growth and final completeness because more informative predicted viewpoints become reachable within one planning step.
This highlights a practical limit of moving object reconstruction: prediction can identify useful future viewpoints, but reconstruction improves only when the robot can physically reach them before the object state changes again.

\textbf{Ablation of belief-space evaluation.}
Figure~\ref{fig:mc_coverage} isolates the effect of viewpoint evaluation under the predictive belief.
Random feasible selection performs worst, showing that the candidate set alone is not sufficient.
Scoring only at the predicted mean object state improves performance, but Monte Carlo expected scoring improves completeness further by accounting for nonlinear changes in viewpoint utility across plausible execution-time object states.
The improvement from $N=10$ to $N=40$ indicates that better approximating the predictive expectation leads to more reliable viewpoint choices.
Average scoring time remains subsecond, increasing modestly from $427\,\mathrm{ms}$ for predicted-mean scoring to $479\,\mathrm{ms}$ for MC scoring with $N=40$.
 

\begin{table}[ht] 
    \caption{\textbf{Ablation of candidate viewpoint generation geometry.}
    Final completeness at iteration 10 for an isotropic ring and the proposed uncertainty-adaptive ellipse.
    The ring radius is determined from the maximum eigenvalue of the predicted object-position covariance. Results are reported as mean $\pm$ std over 10 seeds under three different scoring methods.} 
    \vspace{-8pt}
    \centering  
    \begin{tabular}{lccc}
      \toprule
      Candidate Shape & Random & Predicted Mean & MC\,=\,40 \\
      \midrule
      Ring (isotropic)      & 61.1\% \scriptsize{$\pm$\,8.2\%} & 65.9\% \scriptsize{$\pm$\,9.1\%} & 69.4\% \scriptsize{$\pm$\,10.7\%} \\
      Ellipse (adaptive)    & 64.8\% \scriptsize{$\pm$\,6.4\%} & 67.1\% \scriptsize{$\pm$\,9.5\%} & \textbf{70.8\%} \scriptsize{$\pm$\,8.7\%} \\ 
      \bottomrule
    \end{tabular}  
  \label{tab:ring_ellipse}  
  \vspace{-0.15 in}
\end{table}


\textbf{Ablation of candidate geometry.}
Table~\ref{tab:ring_ellipse} isolates the effect of uncertainty-aware candidate viewpoint generation.
The adaptive ellipse improves final completeness across all three scoring methods, indicating that directional uncertainty helps place candidate viewpoints in regions that remain useful under plausible object motion.
The gains are smaller than those from Monte Carlo belief-space evaluation, indicating that candidate generation and expected scoring are complementary, with belief-space evaluation contributing the larger gain.

\subsection{Real-world Experiments}
\label{sec:real-world}

The real-world experiments test whether the simulation trends transfer to a physical robot with real depth sensing and injected object-position measurement noise.
A mobile robot is equipped with an Intel RealSense D435i depth camera, as shown in Fig.~\ref{fig:platform}. 
The moving object follows a predefined S-curve trajectory while the robot replans viewpoints for 10 iterations.
A motion-capture system provides ground-truth object poses for evaluation.
Noisy planar position measurements are generated by adding zero-mean Gaussian noise to the measured object position, with $\sigma\in\{0.10,0.15,0.20\}\,\mathrm{m}$.
For each setting, four trials are conducted using the same three viewpoint selection methods as in Sec.~\ref{sec:sim}.
Figure~\ref{fig:real_experiment} shows one real-world trial.
The robot changes its relative viewpoint over successive planning iterations rather than simply following the object.
This behavior is important for reconstruction: as the object moves, the planner selects views that expose new object surfaces instead of only maintaining proximity.

\textbf{Baseline comparison under measurement noise.}
Figure~\ref{fig:real_progress} compares the three viewpoint selection methods as injected object-position measurement noise increases.
Predictive NBV achieves the highest final completeness across all noise settings.
The trend matches simulation: Predictive NBV continues to improve over later iterations, while Non-predictive NBV and Tracking-only often plateau earlier.
This indicates that views selected from the current belief or by proximity to the predicted target can become less informative after the object moves.
Because the methods share the same smoothing module, reachability constraints, and reconstruction update, the performance gap reflects the viewpoint-selection strategy: whether prediction is propagated to execution time and whether candidate views are scored for coverage.
The results show that motion-uncertainty-aware NBV transfers to the physical robot and improves reconstruction when noisy partial observations induce uncertainty over the object state.

\section{Limitations}

Our framework has the following limitations.
First, viewpoint evaluation and voxel integration rely on accurate camera extrinsics and robot-to-object transforms over the planning horizon. Localization drift, synchronization error, or calibration bias can distort the predicted camera-object configuration and degrade viewpoint scoring. Future work could incorporate robot pose uncertainty and online calibration checks into the NBV planning pipeline. 

Second, the NBV planner relies on externally provided noisy object-position measurements at each planning time. 
This separates active reconstruction from the upstream problem of onboard target detection and tracking. 
Performance may degrade when target observations are intermittent, biased, or delayed. 
A natural extension is to integrate onboard target tracking with reconstruction, so that the robot jointly estimates the object state and selects viewpoints that support both tracking and surface coverage.

Third, NBV evaluation is performed on an object-centric bounded voxel proxy using a projection-based surrogate score. This introduces approximations from voxel discretization, frontier extraction, ellipsoid fitting, and simplified depth-ordering rather than explicit ray casting. These approximations may become brittle for thin structures, strong self-occlusion, or degraded depth measurements.

\section{Conclusions}

This work presented a motion-uncertainty-aware next-best-view (NBV) planning framework for active reconstruction of a moving object. 
The framework addresses a key limitation of standard NBV planning for static objects: when the object moves, the viewpoint selected during planning may no longer be informative at execution time. 
Instead of evaluating candidate views at a single predicted object pose, the proposed method propagates a predictive belief over the plausible object state and selects viewpoints by maximizing the expected coverage-driven score over that belief.

The resulting online planner integrates fixed-lag Gaussian Process prediction, uncertainty-adaptive candidate generation, reachability filtering, and Monte Carlo belief-space evaluation. 
Simulation and real-world experiments show that this formulation improves reconstruction completeness over both non-predictive NBV and prediction-only tracking baselines. 
The baseline comparisons show that motion prediction alone is not sufficient, and coverage-driven NBV without execution-time prediction is also insufficient under object motion uncertainty. 
Together, these results demonstrate that moving-object reconstruction benefits from explicitly coupling predictive state estimation with coverage-driven viewpoint selection.


\section{Acknowledgment}
This study was funded by Project CETI via grants from Dalio Philanthropies and Ocean X; Sea Grape Foundation; Virgin Unite and Rosamund Zander/Hansjorg Wyss through The Audacious Project: a collaborative funding initiative housed at TED. 
This work has been made possible in part by a gift from the Chan Zuckerberg Initiative Foundation to establish the Kempner Institute for the Study of Natural and Artificial Intelligence.


\bibliographystyle{IEEEtran}
\bibliography{references} 

\end{document}